\documentclass[11pt]{article}
\usepackage[T1]{fontenc}
\usepackage{lmodern}
\usepackage[a4paper,margin=25mm]{geometry}
\usepackage{microtype,seqsplit}
\usepackage{amsmath,amssymb}
\usepackage{graphicx,booktabs,longtable,array,tabularx}
\usepackage[numbers,sort&compress]{natbib}
\usepackage{caption}
\usepackage[section]{placeins}
\usepackage{flafter}
\usepackage{xurl}
\usepackage{xcolor}
\usepackage[colorlinks=true,linkcolor=blue!50!black,citecolor=blue!50!black,urlcolor=blue!50!black]{hyperref}
\usepackage{fancyhdr}
\hypersetup{pdftitle={Decision Models: An Introduction and Capability Evaluation},pdfauthor={Rongtao Zhang, Mingjing Qi}}
\newcommand{\fig}[3]{\begin{figure}[!htbp]\centering\includegraphics[width=\linewidth]{figures/#1.pdf}\caption{#2}\label{#3}\end{figure}}

\title{Decision Models: An Introduction and Capability Evaluation}
\author{Rongtao Zhang, Mingjing Qi\\
School of Artificial Intelligence, Beihang University\\
\texttt{rongtaozhang@buaa.edu.cn}\quad\texttt{mingjingqi@buaa.edu.cn}}
\date{November 2026}
\begin{document}
\maketitle
\begin{abstract}
Applications such as ticket routing and incident triage need decisions that software can act on. Decision models provide these outputs through explicit types for choices, binary judgments, and policy scores. We evaluate the accuracy, output validity, and request latency of eleven configurations on 1,000 English and Chinese cases across nine task suites. StartLux Instant achieves 95.3\% overall accuracy, returns valid outputs on all 1,000 cases, and has a median request latency of 0.084\,s. It reaches 99.4\% on classification and routing and selects the correct response on all 100 tool-selection cases. The highest overall accuracy among the evaluated configurations is 97.1\%. These results show how a typed decision service can support fast, programmatic decisions and why its suitability depends on task-specific errors and latency requirements.
\end{abstract}
\section{Introduction}
An application that routes a support ticket, checks whether a passage supports a statement, or assigns an incident severity needs an answer that can be consumed by a program. The operational output is often a member of a finite set, a Boolean value, or a number interpreted under a documented policy. A \emph{decision model} exposes an interface that maps an application state and an explicit question schema to a typed decision.

A typed decision gives downstream software a defined result to validate and act on. A router can dispatch a ticket using the returned queue identifier, and a policy engine can use a severity grade to decide whether to escalate an incident. Specialized decision models and general generative models can both expose this interface. Its value in either case depends on how often the output is valid, how often the decision is correct, and how quickly the application receives it.

We evaluate these properties on tasks that involve routing, evidence assessment, response selection, and rule application. Overall accuracy describes performance across the benchmark. Task-level accuracy and rule-score error show where that aggregate is informative for a particular application, while median and tail latency describe the time needed to return a decision.

All eleven configurations answer the same 1,000 cases. The evaluation quantifies where typed decisions can serve as an input to software workflows and where their errors or latency may limit that use.

\section{Typed decision interfaces}
\label{sec:interfaces}
Let $x$ denote the application state and $q$ a question with instructions and an output schema. A decision interface returns $f(x,q)$ subject to a declared type. In this evaluation, every request contains a single question. Three primitives are used:
\begin{itemize}
\item \texttt{choice}: return an identifier from the supplied candidate set. The benchmark uses between 2 and 20 candidates; exact identifier agreement determines correctness.
\item \texttt{noul}: return a number in $[0,1]$ interpreted as a yes/no decision at a threshold of 0.5. Generative adapters first require an actual JSON Boolean and convert it to 0 or 1.
\item \texttt{score}: return a finite number in $[0,3]$ for the supplied four-level rubric. Fractional predictions are permitted and evaluated by absolute error.
\end{itemize}
A legal output can still be incorrect: a valid option may be wrong, and a well-formed number may violate the policy. Decision quality therefore requires measures of answer correctness, output validity, and score error.

\paragraph{Direct readout.}
StartLux Instant reads probabilities from option-token logits without autoregressive answer generation and uses bidirectional prompt attention in its full-attention layers \citep{startluxinstant,startlux2026decision}. Its decision wrapper exposes these probabilities as typed answers. Clef uses a separate joint schema head to score the options of typed questions from backbone states \citep{clef}. These implementations illustrate two ways to provide direct typed output.

\paragraph{Conditional deliberation.}
StartLux AutoThink produces an initial decision readout. A gate can trigger a generated thought, after which the model combines the initial and updated readouts \citep{startluxauto}. The service returns the same typed decision in either case. Table~\ref{tab:config} lists the two AutoThink configurations, and Appendix~\ref{app:configs} specifies their controls.

\paragraph{Structured output in other services.}
Qwen3.5-27B and Qwen3.8-27B generate answers in the required JSON format \citep{qwen35,qwen38}. GPT returns structured answers through the Codex app-server. JEV provides a hosted native typed-output service. Table~\ref{tab:config} lists the evaluated configurations, and Appendix~\ref{app:configs} documents their settings.

Under the benchmark incident policy, when a core workflow is blocked for all users and there is no data loss, the severity grade is 3 without a workaround and 1 with a workaround.

\section{Evaluation benchmark}
\label{sec:data}
\subsection{Composition}
The benchmark contains 1,000 cases across nine task suites: 900 text-only cases and 100 single-image cases. It includes 700 English and 300 Chinese cases, with language labels also covering mixed-language source material or answer options. The output types are \texttt{choice} for 816 cases, \texttt{noul} for 134, and \texttt{score} for 50.

Table~\ref{tab:dataset} gives the task allocation and source resources. The benchmark uses curated subsets and adaptations of these resources. Suite sizes determine their contribution to overall accuracy: classification, semantics, and document evidence each contain 160 cases, while numerical reasoning contains 50.

\begin{table}[!htbp]
\centering\small
\caption{The 1,000-case benchmark: task coverage, case counts, and source resources. Source splits are listed in Appendix~\ref{app:sources}.}
\label{tab:dataset}
\begin{tabularx}{\linewidth}{llrX}
\toprule
Suite & Task & Cases & Sources \\
\midrule
S01 & Classification and routing & 160 & BANKING77 (80), TNEWS (80) \\
S02 & Semantic relations & 160 & OCNLI (100), PAWS-X (60) \\
S03 & Document evidence & 160 & BoolQ (60), C3 dialogue (30), C3 long (30), ContractNLI (40) \\
S04 & Product relevance & 150 & ESCI (150) \\
S05 & Tool-response selection & 100 & When2Call multiple-choice subset (100) \\
S06 & Numerical reasoning & 50 & AQuA (50) \\
S07 & Perturbation robustness & 70 & DecisionRobust: 35 clean--attack pairs \\
S08 & Single-image reasoning & 100 & AI2D (40), ScienceQA (40), MMMU (20) \\
S09 & Rule scoring & 50 & Explicit incident-severity policy: 25 minimal pairs \\
\midrule
 & Total & 1,000 & Nine task suites \\
\bottomrule
\end{tabularx}
\end{table}

The sources cover banking intent detection \citep{casanueva-etal-2020-efficient}, Chinese classification and inference \citep{clue,ocnli}, paraphrase identification \citep{yang-etal-2019-paws}, evidence judgments \citep{clark-etal-2019-boolq,c3,koreeda-manning-2021-contractnli-dataset}, search relevance \citep{esci}, tool-response choice \citep{when2call}, mathematical multiple-choice questions \citep{aqua}, robustness \citep{decisionrobust}, and visual reasoning \citep{ai2d,scienceqa,mmmu}. Appendix~\ref{app:sources} lists resource links and split identities.

\subsection{Common inputs and case grouping}
Each case specifies a state, a question, candidate descriptions or a scoring rubric, and a reference answer. Every configuration receives the same semantic content and legal answer set. Native APIs accept typed questions; generative adapters request the corresponding JSON through their model-specific wrapper, chat template, tokenizer, and image processor.

The case set and reference labels are fixed across configurations. Labels were checked against source answers with AI-assisted review. Robustness items are curated challenge cases.

The 1,000 cases form 940 source-linked groups. Related items, including clean--attack pairs and rule minimal pairs, share group identifiers. Grouped resampling preserves these relationships when estimating uncertainty.

Each visual case supplies one RGB PNG with at most 1,048,576 pixels and no benchmark-side cropping. All configurations receive the same image, with at most one image per request, and apply their native image decoding and vision preprocessing. The 40 AI2D cases balance four answer positions at ten each; the 20 MMMU cases cover two subject subsets at ten each.

The ESCI subset contains 40 Exact, 40 Substitute, 40 Irrelevant, and 30 Complement judgments. The When2Call task measures offline selection of the appropriate proposed response or tool-related action.

\subsection{Rule scoring and robustness}
The 50 rule cases are organized into 25 minimal pairs, each designed to change a decision-relevant fact. The four-level policy is applied in priority order:
\begin{enumerate}
\item Confirmed data loss implies severity 3.
\item A blocked core operation without a workaround implies 3 when all users are affected, and 2 otherwise.
\item A blocked operation with a workaround, or nonblocking degradation, implies 1.
\item All remaining cases imply 0.
\end{enumerate}
These synthetic cases test whether predictions follow the policy when a decision-relevant fact changes.

The robustness suite contains 35 clean--attack pairs: 15 character-perturbation pairs and 20 instruction-injection pairs, with one attack per clean item. We report accuracy on each side and the fraction of pairs for which both answers are correct. Pair accuracy requires a correct decision on both the clean item and its attacked variant.

\section{Models and evaluation settings}
\label{sec:protocol}
\subsection{Model and service configurations}
Table~\ref{tab:config} lists the eleven configurations, their abbreviations, and their decision paths. The local models have nominal 27B parameter identities and differ in output heads, reasoning policies, and serving engines.

\begin{table}[!htbp]
\centering\small
\caption{Eleven tested configurations. ``GPT'' denotes the service identifier \texttt{gpt-6-sol}. Full repositories, revisions, and controls appear in Appendix~\ref{app:configs}.}
\label{tab:config}
\begin{tabularx}{\linewidth}{llX}
\toprule
Abbreviation & Configuration & Decision path and reasoning control \\
\midrule
SL-I & StartLux Instant & Native readout; thinking disabled \\
SL-A & StartLux AutoThink & Native readout; gated thought, gate 0.9 \\
SL-N & StartLux AutoThink (never) & Same AutoThink checkpoint; thought disabled \\
Clef & Clef 27B & Native joint schema head; no answer generation \\
Q35-F & Qwen3.5 fast & Generated JSON; thinking disabled \\
Q35-T & Qwen3.5 thinking & Generated JSON; thinking enabled \\
GPT-M & GPT medium & Hosted service; medium reasoning, no tools \\
GPT-N & GPT none & Hosted service; reasoning control set to none, no tools \\
JEV & JEV 1.14.0 & Hosted typed-output service \\
Q38-F & Qwen3.8 fast & Generated JSON; thinking disabled \\
Q38-M & Qwen3.8 medium & Generated JSON; thinking enabled, medium effort \\
\bottomrule
\end{tabularx}
\end{table}

Local inference uses one NVIDIA RTX PRO 6000 Blackwell GPU with 96\,GB memory, tensor parallelism 1, BF16 weights, a 16,384-token context, and a serving memory target of 0.8. StartLux and Qwen use vLLM 0.31; Clef uses SGLang 0.5.23. Appendix~\ref{app:configs} specifies model revisions, generation controls, and serving settings.

\subsection{Request policy and timing}
Each configuration answers every case once, in the same case order, with one request in flight. Failed and invalid responses count as errors; requests are not retried or replaced. Each configuration's generation controls and output budget are specified in Appendix~\ref{app:configs}.

Latency is the observed request wait, including the service call and response validation. Hosted calls can include network and queue delays; GPT waits also include ephemeral thread setup, the turn, and cleanup. Model loading, payload construction outside the timed request, logging, and diagnostic requests are excluded. Cumulative wait is the sum over all 1,000 requests at concurrency one.

\section{Metrics and uncertainty}
\label{sec:metrics}
\subsection{Decision accuracy and score error}
Discrete accuracy measures correctness on choice and binary questions. A choice is valid only if it is an allowed string identifier; a native binary value must be finite and in $[0,1]$, with values at least 0.5 interpreted as positive. Failed calls, invalid types, truncated answers, and malformed outputs count as incorrect. For the 950 discrete cases $D$,
\begin{equation}
 A_{950}=\frac{1}{950}\sum_{i\in D}\mathbf{1}\{\text{valid}_i\land\hat y_i=y_i\}.
\end{equation}
For a rule case, a finite score $s_i\in[0,3]$ has absolute error $|s_i-y_i|$. An invalid return receives error 3, the full scale range. Mean absolute error over the 50 rule cases $R$ is
\begin{equation}
 \operatorname{MAE}_{50}=\frac{1}{50}\sum_{i\in R}
 \begin{cases}|s_i-y_i|,&\text{valid}_i,\\3,&\text{otherwise}.\end{cases}
\end{equation}

\subsection{Overall accuracy and task diagnostics}
Overall accuracy gives each of the 1,000 cases equal weight. Discrete decisions use exact correctness; a valid rule score is correct when its nearest grade satisfies $\lfloor s_i+0.5\rfloor=y_i$. Thus,
\begin{equation}
 A_{1000}=\frac{950A_{950}+\sum_{i\in R}\mathbf{1}\{\text{valid}_i\land\lfloor s_i+0.5\rfloor=y_i\}}{1000}.
 \label{eq:composite}
\end{equation}
Rule cases contribute 5\% of overall accuracy. This aggregate measures decision and grade correctness, while rule-score MAE measures numerical precision. Task accuracy, text-only and image-only accuracy, accuracy excluding robustness, the unweighted macro-average of the eight discrete suites, invalid counts, and clean--attack pair outcomes provide complementary diagnostics.

\subsection{Stratified grouped bootstrap}
The bootstrap estimates uncertainty due to case sampling. Each of 4,000 replicates samples groups with replacement within each suite using seed 20261111. Every configuration uses the same draws. For suite $s$ with $n_s$ cases and sampled groups $g^*_{s,1},\ldots,g^*_{s,G_s}$, the overall accuracy for model $m$ in replicate $b$ is
\begin{equation}
 \widehat A_m^{(b)}=\sum_{s=1}^{9}\frac{n_s}{1000}
 \frac{\sum_{j=1}^{G_s}c_{m,g^*_{s,j}}}{\sum_{j=1}^{G_s}n_{g^*_{s,j}}},
\end{equation}
where $c_{m,g}$ and $n_g$ are correct and total cases in group $g$. The ratio accommodates unequal group sizes while preserving fixed suite weights. Discrete accuracy uses the eight public suites and denominator 950; score MAE resamples the 25 rule groups. Reported 95\% intervals are the 2.5th and 97.5th percentiles.

For models A and B, the paired difference in replicate $b$ is $100(\widehat A_A^{(b)}-\widehat A_B^{(b)})$ percentage points (pp). We also count the cases answered correctly by only one of the two models. The 55 comparisons are exploratory and use unadjusted intervals. An interval spanning zero leaves the direction of the difference unresolved.

\section{Results}
\label{sec:results}
\FloatBarrier
\subsection{Accuracy, score error, and invalid returns}
GPT medium achieves the highest overall accuracy, 97.1\%, followed by Qwen3.8 medium at 96.5\%, StartLux AutoThink at 95.9\%, and StartLux Instant at 95.3\% (Table~\ref{tab:overall}). Figure~\ref{fig:accuracy} compares all eleven configurations with grouped bootstrap intervals.

\begin{table}[!htbp]\centering\small\setlength{\tabcolsep}{4pt}\caption{Overall accuracy on the 1,000-case benchmark. Discrete and Rule report correct counts for each question type. A rule score is rounded to the nearest grade. Invalid counts cover all cases.}\label{tab:overall}\resizebox{\linewidth}{!}{\begin{tabular}{lrrrrrr}
\toprule
Configuration & Correct & Acc. (\%) & 95\% interval & Discrete & Rule & Invalid \\
\midrule
GPT medium & 971 & 97.10 & [96.10, 98.00] & 921/950 & 50/50 & 0 \\
Qwen3.8 medium & 965 & 96.50 & [95.40, 97.60] & 915/950 & 50/50 & 0 \\
StartLux AutoThink & 959 & 95.90 & [94.60, 97.10] & 909/950 & 50/50 & 0 \\
StartLux Instant & 953 & 95.30 & [94.00, 96.50] & 903/950 & 50/50 & 0 \\
StartLux AutoThink (never) & 944 & 94.40 & [93.00, 95.70] & 894/950 & 50/50 & 0 \\
JEV 1.14.0 & 938 & 93.80 & [92.40, 95.20] & 888/950 & 50/50 & 0 \\
GPT none & 937 & 93.70 & [92.20, 95.10] & 887/950 & 50/50 & 0 \\
Qwen3.5 thinking & 926 & 92.60 & [91.00, 94.20] & 877/950 & 49/50 & 11 \\
Clef 27B & 908 & 90.80 & [89.00, 92.40] & 858/950 & 50/50 & 0 \\
Qwen3.5 fast & 796 & 79.60 & [77.30, 81.90] & 747/950 & 49/50 & 0 \\
Qwen3.8 fast & 782 & 78.20 & [75.80, 80.50] & 734/950 & 48/50 & 0 \\
\bottomrule
\end{tabular}
}\end{table}
\fig{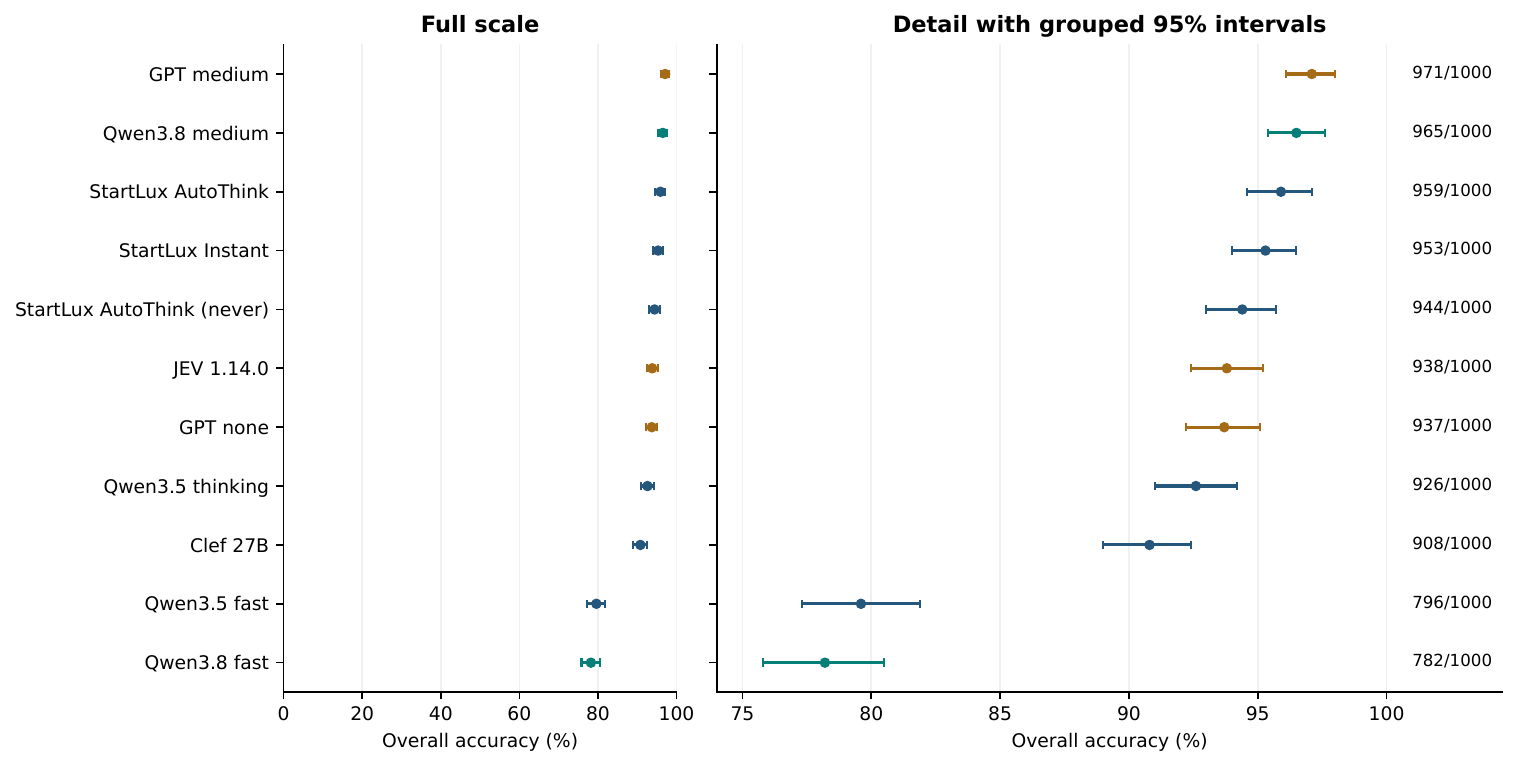}{Overall accuracy with 95\% stratified grouped bootstrap intervals. The left panel shows the full scale; the right panel enlarges the observed range. Colors distinguish StartLux, Clef, and Qwen3.5 (blue), Qwen3.8 (teal), and hosted services (ochre).}{fig:accuracy}

Qwen3.5 thinking produces eleven invalid responses due to output-budget exhaustion or truncation. These responses count as errors. The other ten configurations return valid answers on every case.

The discrete cases yield the same ordering among the four highest-scoring configurations (Table~\ref{tab:primary}). GPT medium answers 921 of 950 correctly (96.95\%), followed by Qwen3.8 medium at 915/950 (96.32\%), StartLux AutoThink at 909/950 (95.68\%), and StartLux Instant at 903/950 (95.05\%). On the 50 rule cases, GPT medium, GPT none, and Qwen3.8 medium have zero observed mean absolute error (MAE). Clef assigns the correct grade to all 50 cases after rounding, with an MAE of 0.0955 on the raw scores.

\begin{table}[!htbp]\centering\small\setlength{\tabcolsep}{4pt}\caption{Discrete-decision accuracy and rule-score MAE. Accuracy intervals are in percent; MAE and its interval are on the 0--3 scale.}\label{tab:primary}\resizebox{\linewidth}{!}{\begin{tabular}{lrrrrr}
\toprule
Configuration & Correct/950 & Acc. (\%) & 95\% interval & Rule MAE & MAE 95\% interval \\
\midrule
StartLux Instant & 903 & 95.05 & [93.68, 96.32] & 0.0004 & [0.0003, 0.0005] \\
StartLux AutoThink & 909 & 95.68 & [94.32, 96.95] & 0.0030 & [0.0024, 0.0036] \\
StartLux AutoThink (never) & 894 & 94.11 & [92.63, 95.47] & 0.0030 & [0.0024, 0.0036] \\
Clef 27B & 858 & 90.32 & [88.42, 92.00] & 0.0955 & [0.0766, 0.1156] \\
Qwen3.5 fast & 747 & 78.63 & [76.21, 81.05] & 0.0400 & [0.0000, 0.1200] \\
Qwen3.5 thinking & 877 & 92.32 & [90.63, 94.00] & 0.0600 & [0.0000, 0.1800] \\
GPT medium & 921 & 96.95 & [95.89, 97.89] & 0.0000 & [0.0000, 0.0000] \\
GPT none & 887 & 93.37 & [91.79, 94.84] & 0.0000 & [0.0000, 0.0000] \\
JEV 1.14.0 & 888 & 93.47 & [92.00, 94.95] & 0.0146 & [0.0122, 0.0172] \\
Qwen3.8 fast & 734 & 77.26 & [74.74, 79.68] & 0.0400 & [0.0000, 0.1000] \\
Qwen3.8 medium & 915 & 96.32 & [95.16, 97.47] & 0.0000 & [0.0000, 0.0000] \\
\bottomrule
\end{tabular}
}\end{table}

\FloatBarrier
\subsection{Paired accuracy across configurations}
Table~\ref{tab:selected} and Figure~\ref{fig:paired} show how the tested configurations differ in correctness on the same cases. Qwen3.8 medium gains 18.3 pp over Qwen3.8 fast, with interval $[16.0,20.8]$. The medium configuration alone is correct on 196 cases, and the fast configuration alone is correct on 13.

\begin{table}[!htbp]\centering\small\setlength{\tabcolsep}{4pt}\caption{Selected paired accuracy comparisons. Difference is A minus B in percentage points. ``A only'' and ``B only'' count discordant cases. Appendix~\ref{app:allpairs} includes every pair.}\label{tab:selected}\resizebox{\linewidth}{!}{\begin{tabular}{llrrrr}
\toprule
A & B & A only & B only & Difference (pp) & 95\% interval (pp) \\
\midrule
Q38-M & Q38-F & 196 & 13 & +18.30 & [+16.00, +20.80] \\
Q38-M & Q35-T & 52 & 13 & +3.90 & [+2.30, +5.50] \\
Q38-F & Q35-F & 75 & 89 & -1.40 & [-3.80, +1.00] \\
Q38-M & GPT-M & 17 & 23 & -0.60 & [-1.80, +0.60] \\
Q38-M & SL-A & 31 & 25 & +0.60 & [-0.90, +2.10] \\
SL-A & SL-N & 21 & 6 & +1.50 & [+0.60, +2.50] \\
GPT-M & GPT-N & 46 & 12 & +3.40 & [+2.00, +4.90] \\
SL-I & JEV & 43 & 28 & +1.50 & [-0.20, +3.00] \\
\bottomrule
\end{tabular}
}\end{table}
\fig{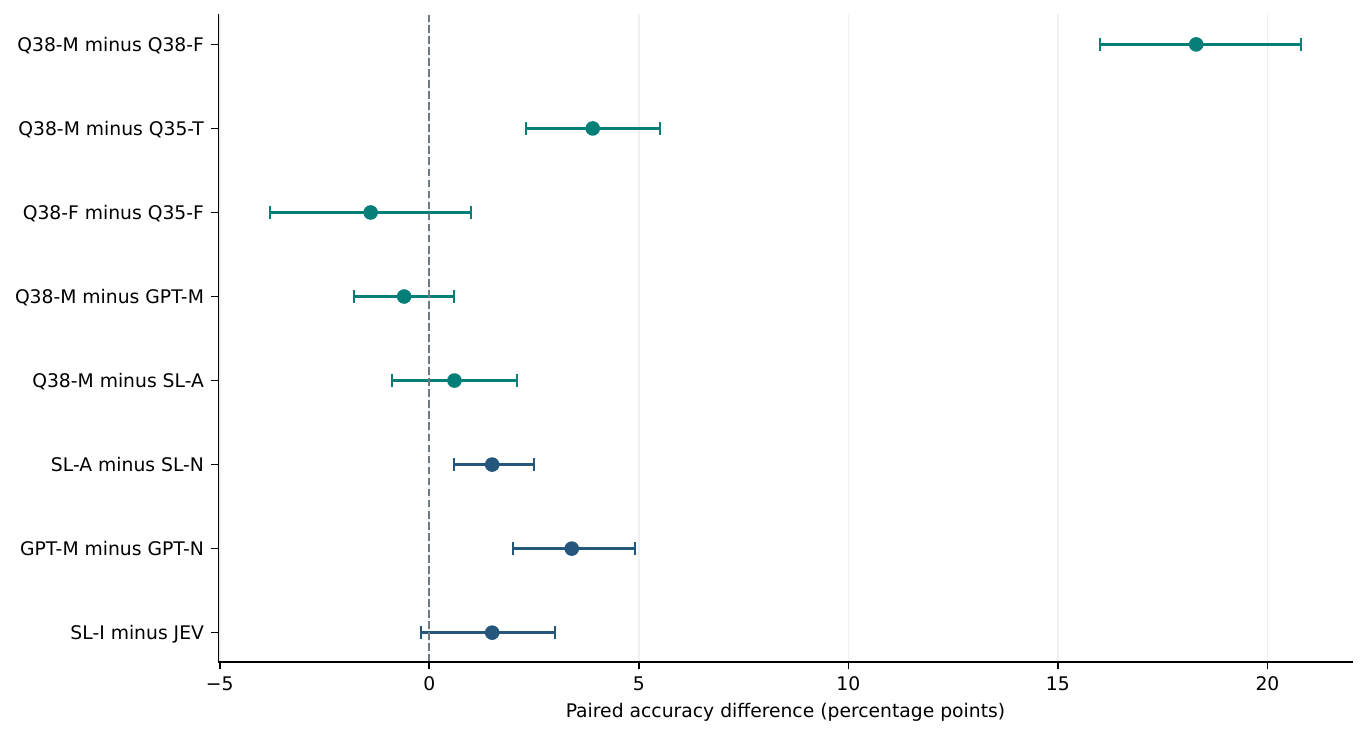}{Selected paired accuracy differences with 95\% bootstrap intervals. Shared resampling preserves the case pairing.}{fig:paired}

Qwen3.8 medium exceeds Qwen3.5 thinking by 3.9 pp, with interval $[2.3,5.5]$. Qwen3.8 fast is 1.4 pp below Qwen3.5 fast, with interval $[-3.8,1.0]$.

Qwen3.8 medium scores $-0.6$ pp relative to GPT medium ($[-1.8,0.6]$), with 17 cases correct only for Qwen3.8 medium and 23 correct only for GPT medium. Its difference from StartLux AutoThink is $+0.6$ pp ($[-0.9,2.1]$). Both intervals span zero, so the direction of each difference remains unresolved under grouped resampling.

On the same AutoThink checkpoint, enabling the gated thought policy raises overall accuracy from 94.4\% to 95.9\%: 21 cases improve and six regress, a net 1.5 pp with interval $[0.6,2.5]$. AutoThink uses its deliberation path on 359/1,000 cases (35.9\%), generating 54,918 thought tokens in total. GPT medium exceeds GPT none by 3.4 pp, with additional request time and reported reasoning tokens.

\FloatBarrier
\subsection{Task structure, rule adherence, and robustness}
Task-level accuracy shows where a decision service is most effective (Figure~\ref{fig:tasks} and Table~\ref{tab:tasks}). StartLux Instant reaches 99.4\% on classification and routing, 98.1\% on document evidence, and 100.0\% on tool-response selection. StartLux AutoThink reaches 96.0\% on numerical reasoning and 94.0\% on single-image questions. Its never configuration scores 78.0\% and 88.0\% on those tasks, respectively, and 92.9\% on robustness compared with 91.4\% for AutoThink. Qwen3.8 fast scores 92.5\%, 26.0\%, and 61.0\% on classification and routing, numerical reasoning, and single-image questions. Qwen3.8 medium scores 96.9\%, 100.0\%, and 95.0\% on the same suites.

\fig{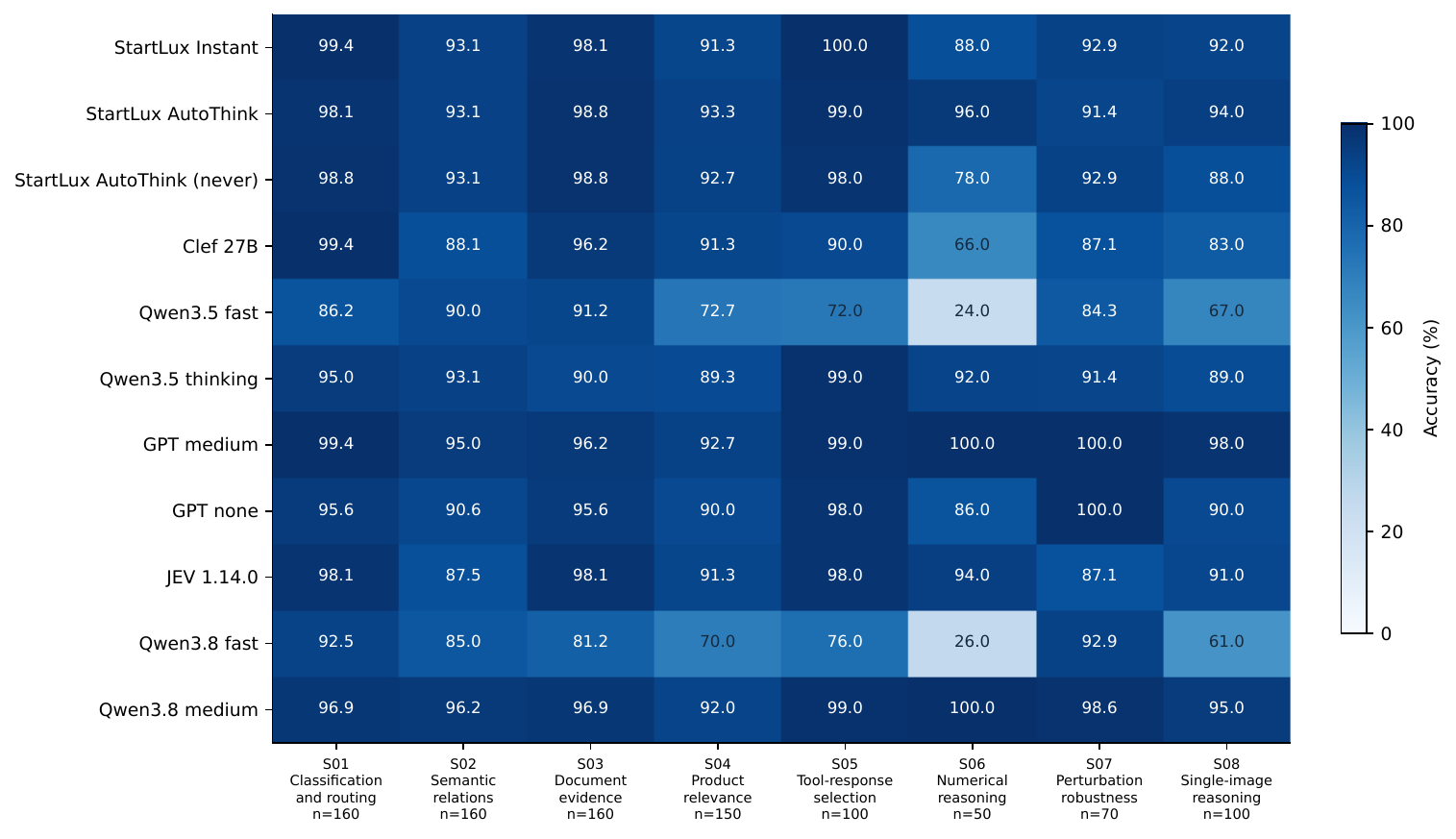}{Accuracy on the eight discrete task suites. The different column sizes define the micro-average; an unweighted macro-average is supplied in Appendix~\ref{app:sensitivity}.}{fig:tasks}
\begin{table}[!htbp]\centering\small\setlength{\tabcolsep}{4pt}\caption{Per-suite discrete accuracy (\%). Suite definitions and sizes are in Table~\ref{tab:dataset}.}\label{tab:tasks}\resizebox{\linewidth}{!}{\begin{tabular}{lrrrrrrrr}
\toprule
Configuration & S01 & S02 & S03 & S04 & S05 & S06 & S07 & S08 \\
\midrule
StartLux Instant & 99.4 & 93.1 & 98.1 & 91.3 & 100.0 & 88.0 & 92.9 & 92.0 \\
StartLux AutoThink & 98.1 & 93.1 & 98.8 & 93.3 & 99.0 & 96.0 & 91.4 & 94.0 \\
StartLux AutoThink (never) & 98.8 & 93.1 & 98.8 & 92.7 & 98.0 & 78.0 & 92.9 & 88.0 \\
Clef 27B & 99.4 & 88.1 & 96.2 & 91.3 & 90.0 & 66.0 & 87.1 & 83.0 \\
Qwen3.5 fast & 86.2 & 90.0 & 91.2 & 72.7 & 72.0 & 24.0 & 84.3 & 67.0 \\
Qwen3.5 thinking & 95.0 & 93.1 & 90.0 & 89.3 & 99.0 & 92.0 & 91.4 & 89.0 \\
GPT medium & 99.4 & 95.0 & 96.2 & 92.7 & 99.0 & 100.0 & 100.0 & 98.0 \\
GPT none & 95.6 & 90.6 & 95.6 & 90.0 & 98.0 & 86.0 & 100.0 & 90.0 \\
JEV 1.14.0 & 98.1 & 87.5 & 98.1 & 91.3 & 98.0 & 94.0 & 87.1 & 91.0 \\
Qwen3.8 fast & 92.5 & 85.0 & 81.2 & 70.0 & 76.0 & 26.0 & 92.9 & 61.0 \\
Qwen3.8 medium & 96.9 & 96.2 & 96.9 & 92.0 & 99.0 & 100.0 & 98.6 & 95.0 \\
\bottomrule
\end{tabular}
}\end{table}

GPT medium and Qwen3.8 medium both answer all 50 numerical cases correctly. GPT medium answers 98/100 image questions correctly; Qwen3.8 medium answers 95, StartLux AutoThink 94, and StartLux Instant 92.

StartLux Instant answers all 35 original items and 30 of 35 perturbed items correctly. Table~\ref{tab:rules} and Figure~\ref{fig:rules} report the full robustness and rule-scoring results.

\begin{table}[!htbp]\centering\small\setlength{\tabcolsep}{4pt}\caption{Rule scoring (50 cases) and robustness (35 pairs, 70 items). Rule counts correct rounded grades. Clean and Attack count correct original and perturbed items. Both counts pairs with two correct answers.}\label{tab:rules}\resizebox{\linewidth}{!}{\begin{tabular}{lrrrrrr}
\toprule
Configuration & MAE & MAE 95\% interval & Rule & Clean & Attack & Both \\
\midrule
StartLux Instant & 0.0004 & [0.0003, 0.0005] & 50/50 & 35/35 & 30/35 & 30/35 \\
StartLux AutoThink & 0.0030 & [0.0024, 0.0036] & 50/50 & 35/35 & 29/35 & 29/35 \\
StartLux AutoThink (never) & 0.0030 & [0.0024, 0.0036] & 50/50 & 35/35 & 30/35 & 30/35 \\
Clef 27B & 0.0955 & [0.0766, 0.1156] & 50/50 & 35/35 & 26/35 & 26/35 \\
Qwen3.5 fast & 0.0400 & [0.0000, 0.1200] & 49/50 & 35/35 & 24/35 & 24/35 \\
Qwen3.5 thinking & 0.0600 & [0.0000, 0.1800] & 49/50 & 34/35 & 30/35 & 29/35 \\
GPT medium & 0.0000 & [0.0000, 0.0000] & 50/50 & 35/35 & 35/35 & 35/35 \\
GPT none & 0.0000 & [0.0000, 0.0000] & 50/50 & 35/35 & 35/35 & 35/35 \\
JEV 1.14.0 & 0.0146 & [0.0122, 0.0172] & 50/50 & 35/35 & 26/35 & 26/35 \\
Qwen3.8 fast & 0.0400 & [0.0000, 0.1000] & 48/50 & 35/35 & 30/35 & 30/35 \\
Qwen3.8 medium & 0.0000 & [0.0000, 0.0000] & 50/50 & 35/35 & 34/35 & 34/35 \\
\bottomrule
\end{tabular}
}\end{table}
\fig{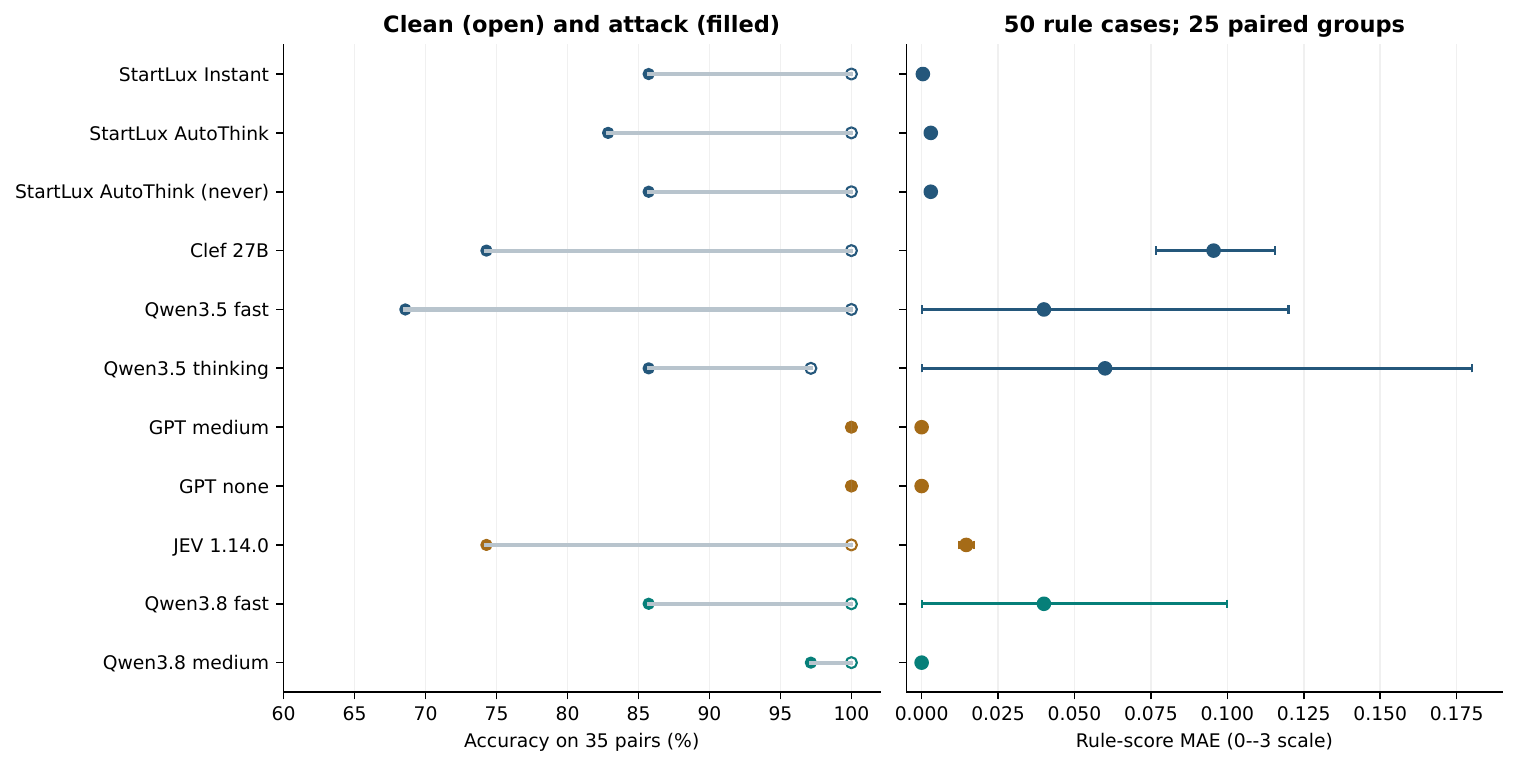}{Left panel shows accuracy on original (open) and perturbed (filled) items from 35 pairs. Lines connect the two observed accuracies. Right panel shows MAE on 50 rule cases with 95\% grouped bootstrap intervals.}{fig:rules}

\FloatBarrier
\subsection{Observed request waits and token accounting}
Table~\ref{tab:latency} and Figure~\ref{fig:latency} report waits over all 1,000 requests, including failures. StartLux AutoThink with thinking disabled has the lowest median, 0.070\,s, followed by StartLux Instant at 0.084\,s and Clef at 0.102\,s. Gated AutoThink has median 0.120\,s but P95 8.605\,s, consistent with a mixture of direct and deliberative requests. Qwen3.8 medium has median 4.866\,s and P95 13.355\,s. Qwen3.5 thinking has median 51.993\,s and P95 94.163\,s.

\begin{table}[!htbp]\centering\small\setlength{\tabcolsep}{4pt}\caption{Request waits over 1,000 cases. Total is cumulative wait, including failed requests.}\label{tab:latency}\resizebox{\linewidth}{!}{\begin{tabular}{lrrrrr}
\toprule
Configuration & P50 (s) & P95 (s) & Max. (s) & Total (min) & Invalid \\
\midrule
StartLux Instant & 0.084 & 0.318 & 1.120 & 2.12 & 0 \\
StartLux AutoThink & 0.120 & 8.605 & 117.962 & 38.76 & 0 \\
StartLux AutoThink (never) & 0.070 & 0.177 & 0.767 & 1.56 & 0 \\
Clef 27B & 0.102 & 0.217 & 0.672 & 1.90 & 0 \\
Qwen3.5 fast & 0.278 & 0.444 & 0.976 & 4.99 & 0 \\
Qwen3.5 thinking & 51.993 & 94.163 & 173.335 & 825.01 & 11 \\
GPT medium & 4.425 & 6.768 & 22.536 & 78.92 & 0 \\
GPT none & 4.029 & 5.837 & 9.884 & 71.07 & 0 \\
JEV 1.14.0 & 0.251 & 0.438 & 2.064 & 4.76 & 0 \\
Qwen3.8 fast & 0.257 & 0.428 & 0.877 & 4.59 & 0 \\
Qwen3.8 medium & 4.866 & 13.355 & 73.933 & 97.39 & 0 \\
\bottomrule
\end{tabular}
}\end{table}
\fig{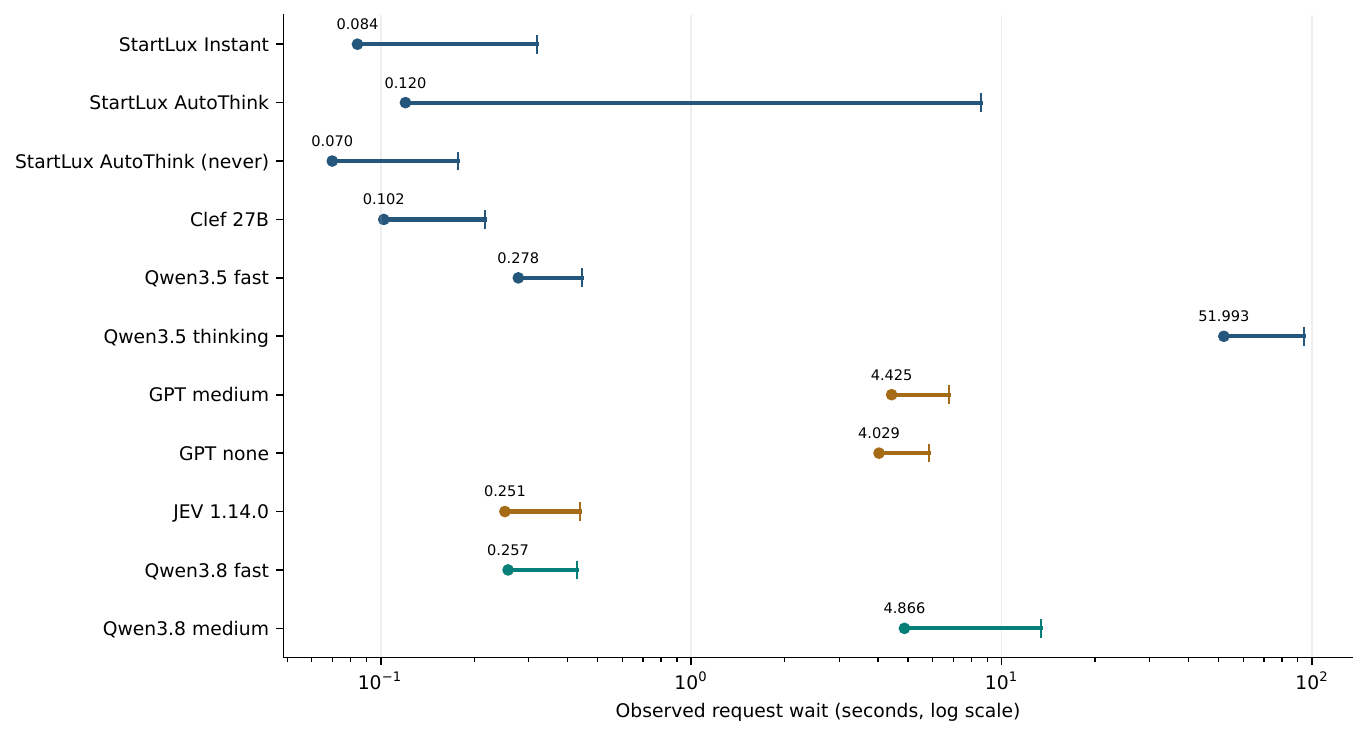}{Median (circle) and P95 (vertical tick) request waits, on a logarithmic scale. Labels above circles give medians in seconds.}{fig:latency}
\fig{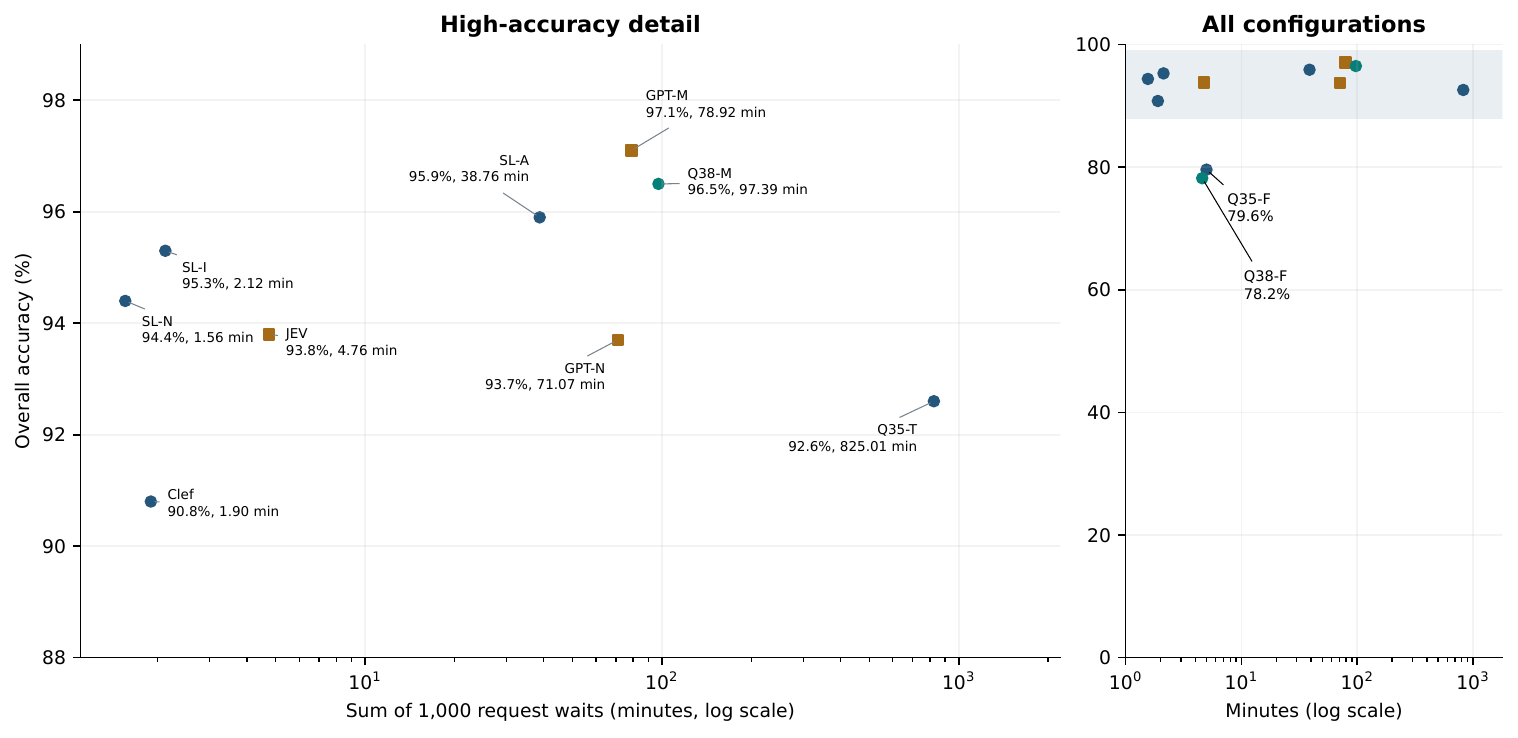}{Overall accuracy against cumulative request wait. The left panel enlarges the high-accuracy region; the right panel shows the full accuracy scale. Labels use the abbreviations in Table~\ref{tab:config}.}{fig:time}

Cumulative waits are 2.12 minutes for StartLux Instant, 38.76 for StartLux AutoThink, 78.92 for GPT medium, 97.39 for Qwen3.8 medium, and 825.01 for Qwen3.5 thinking.

Reported Qwen reasoning tokens total 2,638,718 for Qwen3.5 thinking and 348,386 for Qwen3.8 medium, approximately a 7.6-fold difference. GPT input usage includes the app-server's system context. Appendix~\ref{app:usage} reports all usage fields and distinguishes unavailable reasoning counts from zero.

\section{Discussion}
A routing or triage system can use a typed decision to select its next action immediately after validation. StartLux Instant returns valid outputs on all 1,000 cases and achieves 95.3\% overall accuracy. Its request latency is 0.084\,s at the median and 0.318\,s at the 95th percentile. These measurements characterize the accuracy and request time of one local decision service on the benchmark. For a workflow with a response deadline, tail latency helps determine whether the service fits the available time.

The tool-selection and rule-scoring results show where StartLux Instant can support software workflows. Its errors on perturbed inputs matter for workflows that accept externally supplied content.

The accuracy figures describe the selected cases and their suite weights. Reference labels were checked against source answers with AI assistance. Each accuracy comparison reflects the complete configuration, including its output head, adapter, and generation settings. The Qwen fast and thinking modes vary both reasoning and sampling settings, and the two thinking configurations also differ in presence penalty and effort control. Latency comparisons reflect the serving engines, cache policies, and network paths used for the measured requests.

Table~A.2 in Appendix~A.3 of the StartLux technical report \citep{startlux2026decision} lists the official training splits of BANKING77, ContractNLI, Amazon ESCI, and When2Call as Stage~2 training sources for the preview models. Our benchmark draws 370 of its 1,000 cases from the test splits of these four datasets. Item-level overlap between these test cases and the training data for the final StartLux checkpoints evaluated here remains unverified.

Each case has one response per configuration. Grouped intervals describe sample variation in those responses; repeated inference would measure additional generation and service variation. Independently reviewed deployment samples, matched serving controls, calibration and abstention measures, and interactive tool tasks would extend the evaluation to further application requirements.

\section{Data and reproducibility}
Appendices~\ref{app:configs} and~\ref{app:sources} specify model revisions, execution settings, source URLs, and data splits.

\paragraph{AI assistance.}
AI tools assisted source inspection, benchmark review, analysis scripting, figure preparation, and manuscript drafting.

\section{Conclusion}
Decision models give software a defined output for routing, evidence judgments, response selection, and rule application. On this 1,000-case benchmark, StartLux Instant combines 95.3\% overall accuracy and valid outputs on every case with a median request latency of 0.084\,s. These results show the potential for low-latency decision services in workflows that can consume typed outputs directly. The suitability of a service for a given workflow depends on its error rate for that task, the consequences of an incorrect decision, and the time available to act.

\FloatBarrier
\bibliographystyle{plainnat}
\bibliography{references}
\clearpage
\appendix
\section{Source resources and adapted splits}
\label{app:sources}
Table~\ref{tab:sources} lists source resources, case counts, data splits, and links. The benchmark uses curated subsets and adaptations of these resources; rule-scoring cases follow the policy in Section~\ref{sec:data}.

\begin{longtable}{p{.20\linewidth}rp{.24\linewidth}p{.39\linewidth}}
\caption{Source case counts, data splits, and resource links. All cases are subsets or adaptations.}\label{tab:sources}\\
\toprule Resource & $n$ & Source split & Primary resource \\
\midrule\endfirsthead
\toprule Resource & $n$ & Source split & Primary resource \\
\midrule\endhead
\bottomrule\endfoot
AI2D & 40 & test & \url{https://allenai.org/data/diagrams} \\
AQuA & 50 & test & \url{https://github.com/google-deepmind/AQuA} \\
BANKING77 & 80 & test & \url{https://github.com/PolyAI-LDN/task-specific-datasets} \\
BoolQ & 60 & SuperGLUE validation / BoolQ development & \url{https://github.com/google-research-datasets/boolean-questions} \\
C3 dialogue & 30 & validation & \url{https://github.com/nlpdata/c3} \\
C3 long & 30 & validation & \url{https://github.com/nlpdata/c3} \\
ContractNLI & 40 & test & \url{https://stanfordnlp.github.io/contract-nli/} \\
DecisionRobust & 70 & test & \url{https://huggingface.co/datasets/yflantmy/DecisionRobust} \\
ESCI & 150 & test & \url{https://github.com/amazon-science/esci-data} \\
MMMU & 20 & validation & \url{https://huggingface.co/datasets/MMMU/MMMU} \\
OCNLI & 100 & development & \url{https://github.com/CLUEbenchmark/OCNLI} \\
PAWS-X & 60 & test & \url{https://github.com/google-research-datasets/paws} \\
ScienceQA & 40 & test & \url{https://scienceqa.github.io/} \\
TNEWS & 80 & development & \url{https://github.com/CLUEbenchmark/CLUE} \\
When2Call & 100 & test / mcq & \url{https://huggingface.co/datasets/nvidia/When2Call} \\
Rule grading & 50 & Policy cases & Rubric in Section~\ref{sec:data}; 25 minimal pairs \\
\end{longtable}

\section{Pinned models and execution controls}
\label{app:configs}
The following settings identify the evaluated model versions, reasoning policies, and inference engines.

\paragraph{StartLux Instant (SL-I).}
Repository \url{https://huggingface.co/startlux-models/StartLux-Decision-v1.1-27B-Instant}; revision \texttt{\seqsplit{32e7d35d74db615c4a059181efbf21c5421f3561}}. vLLM 0.31; thinking never; no multi-token prediction (MTP), prefix cache, or chunked prefill; eager execution and FLASH\_ATTN. Required bidirectional prompt attention is enabled.

\paragraph{StartLux AutoThink (SL-A and SL-N).}
Repository \url{https://huggingface.co/startlux-models/StartLux-Decision-v1.1-27B-AutoThink}; revision \texttt{\seqsplit{dd62d5cd84a483d75bdd548880bbf08bdd3f67a4}}. vLLM 0.31; auto versus never thinking; auto gate 0.9, thought budget 4,096, mixture 0.75; no MTP; prefix cache and chunked prefill enabled.

\paragraph{Clef (Clef).}
Repository \url{https://huggingface.co/Cloudflare/clef}; revision \texttt{\seqsplit{cfddda043dd190ef3b01cff449d5ee8cff1e8ec6}}. SGLang 0.5.23, joint schema head, no answer generation, radix cache disabled, CUDA graphs disabled.

\paragraph{Qwen3.5 (Q35-F and Q35-T).}
Repository \url{https://huggingface.co/Qwen/Qwen3.5-27B}; revision \texttt{\seqsplit{fc05daec18b0a78c049392ed2e771dde82bdf654}}. vLLM 0.31 with MTP using two speculative tokens, prefix cache, and chunked prefill. Fast: thinking disabled, temperature 0.7, top-$p$ 0.8, top-$k$ 20, presence penalty 1.5, output budget 256. Thinking: enabled, temperature 1.0, top-$p$ 0.95, top-$k$ 20, presence penalty 1.5, budget 8,192. Thinking mode uses no named effort setting.

\paragraph{Qwen3.8 (Q38-F and Q38-M).}
Repository \url{https://huggingface.co/Qwen/Qwen3.8-27B}; revision \texttt{\seqsplit{1d4bf0f2ff6012fd82039f2fa52739d0dd7c60c0}}. vLLM 0.31, Transformers 5.17, PyTorch 2.13, with two-token MTP, prefix cache, and chunked prefill. Fast uses the same temperature, top-$p$, top-$k$, presence penalty, and budget as Q35-F. Medium: thinking enabled, effort medium, temperature 1.0, top-$p$ 0.95, top-$k$ 20, presence penalty 0, budget 8,192. Both use seed 20261030, min-$p$ 0, repetition penalty 1, and \texttt{preserve\_thinking=false}.

\paragraph{Hosted services (GPT-M, GPT-N, and JEV).}
GPT uses service identifier \texttt{gpt-6-sol} with medium or none reasoning. Each request uses an independent ephemeral Codex app-server thread without tools. System context and thread lifecycle are included in this service path. JEV exposes version \texttt{jev-1.14.0} through a TypeSafe decision interface. Weight revisions, parameter counts, precision, and hardware are not available for these hosted services.

\subsection{Serving conditions}
Measurements were collected in November 2026. Caches are not reset between the two AutoThink modes or between the Qwen3.8 modes; the Qwen3.8 cache state also includes exploratory requests. Qwen3.5 thinking measurements span two serving processes. Qwen3.8 uses the internal architecture class \texttt{Qwen3\_5ForConditionalGeneration}.

\section{Sensitivity and reported usage}
\label{app:sensitivity}
Table~\ref{tab:sensitivity} compares discrete accuracy under alternative task groupings. Text contains the 850 text-only discrete cases; Image contains the 100 single-image cases. ``No robust'' excludes the 70 robustness cases. Macro weights each of the eight discrete suites equally.
\begin{table}[!htbp]\centering\small\setlength{\tabcolsep}{4pt}\caption{Discrete-accuracy sensitivity views (\%). Denominators are shown in the headings. Macro is the unweighted average over eight discrete suites.}\label{tab:sensitivity}\resizebox{\linewidth}{!}{\begin{tabular}{lrrrr}
\toprule
Configuration & Text (850) & Image (100) & No robust (880) & Macro (8) \\
\midrule
StartLux Instant & 95.41 & 92.00 & 95.23 & 94.35 \\
StartLux AutoThink & 95.88 & 94.00 & 96.02 & 95.47 \\
StartLux AutoThink (never) & 94.82 & 88.00 & 94.20 & 92.52 \\
Clef 27B & 91.18 & 83.00 & 90.57 & 87.65 \\
Qwen3.5 fast & 80.00 & 67.00 & 78.18 & 73.43 \\
Qwen3.5 thinking & 92.71 & 89.00 & 92.39 & 92.36 \\
GPT medium & 96.82 & 98.00 & 96.70 & 97.54 \\
GPT none & 93.76 & 90.00 & 92.84 & 93.23 \\
JEV 1.14.0 & 93.76 & 91.00 & 93.98 & 93.15 \\
Qwen3.8 fast & 79.18 & 61.00 & 76.02 & 73.08 \\
Qwen3.8 medium & 96.47 & 95.00 & 96.14 & 96.82 \\
\bottomrule
\end{tabular}
}\end{table}

\label{app:usage}
Table~\ref{tab:usage} gives endpoint-reported input, output, reasoning, and thought usage. A dash denotes an unreported field. Native triggers count cases with a positive native thought field where that metadata is available. GPT input tokens include app-server system context.
\begin{table}[!htbp]\centering\small\setlength{\tabcolsep}{4pt}\caption{Endpoint-reported usage totals over 1,000 cases per configuration. Usage is the number of cases with usage metadata; Native triggers is the number of cases triggering a native thought. Reasoning tokens are shown only where explicitly reported.}\label{tab:usage}\resizebox{\linewidth}{!}{\begin{tabular}{lrrrrr}
\toprule
Configuration & Input tokens & Output tokens & Reasoning tokens & Native triggers & Usage \\
\midrule
StartLux Instant & 342491 & 0 & -- & 0/1000 & 1000/1000 \\
StartLux AutoThink & 342491 & 54918 & -- & 359/1000 & 1000/1000 \\
StartLux AutoThink (never) & 342491 & 0 & -- & 0/1000 & 1000/1000 \\
Clef 27B & 445364 & 0 & -- & -- & 1000/1000 \\
Qwen3.5 fast & 431001 & 8343 & 0 & -- & 1000/1000 \\
Qwen3.5 thinking & 444001 & 2670281 & 2638718 & -- & 1000/1000 \\
GPT medium & 7138572 & 41705 & 24517 & -- & 1000/1000 \\
GPT none & 7123723 & 15589 & 0 & -- & 1000/1000 \\
JEV 1.14.0 & 698439 & 66104 & -- & -- & 1000/1000 \\
Qwen3.8 fast & 431001 & 9374 & 0 & -- & 1000/1000 \\
Qwen3.8 medium & 444001 & 356546 & 348386 & -- & 1000/1000 \\
\bottomrule
\end{tabular}
}\end{table}
\FloatBarrier

\clearpage
\section{All paired accuracy comparisons}
\label{app:allpairs}
Table~\ref{tab:allpairs} gives all 55 unordered configuration pairs. Both, A only, B only, and Neither partition the 1,000 cases. The accuracy difference is $(\text{A only}-\text{B only})/10$ pp. Percentile intervals use the shared stratified grouped draws in Section~\ref{sec:metrics}.
\begingroup\small\setlength{\tabcolsep}{5pt}
\begin{longtable}{llrrrrrr}
\caption{All 55 paired accuracy comparisons on 1,000 cases. Abbreviations are defined in Table~\ref{tab:config}. Intervals are unadjusted.}\label{tab:allpairs}\\
\toprule
A & B & Both & A only & B only & Neither & $\Delta$ (pp) & 95\% interval (pp) \\
\midrule
\endfirsthead
\toprule
A & B & Both & A only & B only & Neither & $\Delta$ (pp) & 95\% interval (pp) \\
\midrule
\endhead
\bottomrule
\endfoot
SL-I & SL-A & 937 & 16 & 22 & 25 & -0.60 & [-1.80, +0.50] \\
SL-I & SL-N & 930 & 23 & 14 & 33 & +0.90 & [-0.30, +2.00] \\
SL-I & Clef & 884 & 69 & 24 & 23 & +4.50 & [+2.70, +6.30] \\
SL-I & Q35-F & 778 & 175 & 18 & 29 & +15.70 & [+13.30, +18.00] \\
SL-I & Q35-T & 892 & 61 & 34 & 13 & +2.70 & [+0.80, +4.50] \\
SL-I & GPT-M & 933 & 20 & 38 & 9 & -1.80 & [-3.20, -0.40] \\
SL-I & GPT-N & 911 & 42 & 26 & 21 & +1.60 & [-0.00, +3.20] \\
SL-I & JEV & 910 & 43 & 28 & 19 & +1.50 & [-0.20, +3.00] \\
SL-I & Q38-F & 764 & 189 & 18 & 29 & +17.10 & [+14.80, +19.60] \\
SL-I & Q38-M & 930 & 23 & 35 & 12 & -1.20 & [-2.70, +0.20] \\
SL-A & SL-N & 938 & 21 & 6 & 35 & +1.50 & [+0.60, +2.50] \\
SL-A & Clef & 892 & 67 & 16 & 25 & +5.10 & [+3.40, +6.80] \\
SL-A & Q35-F & 779 & 180 & 17 & 24 & +16.30 & [+13.90, +18.70] \\
SL-A & Q35-T & 894 & 65 & 32 & 9 & +3.30 & [+1.30, +5.20] \\
SL-A & GPT-M & 937 & 22 & 34 & 7 & -1.20 & [-2.60, +0.20] \\
SL-A & GPT-N & 909 & 50 & 28 & 13 & +2.20 & [+0.50, +4.00] \\
SL-A & JEV & 915 & 44 & 23 & 18 & +2.10 & [+0.50, +3.70] \\
SL-A & Q38-F & 764 & 195 & 18 & 23 & +17.70 & [+15.40, +20.20] \\
SL-A & Q38-M & 934 & 25 & 31 & 10 & -0.60 & [-2.10, +0.90] \\
SL-N & Clef & 889 & 55 & 19 & 37 & +3.60 & [+2.00, +5.20] \\
SL-N & Q35-F & 772 & 172 & 24 & 32 & +14.80 & [+12.40, +17.30] \\
SL-N & Q35-T & 883 & 61 & 43 & 13 & +1.80 & [-0.20, +3.80] \\
SL-N & GPT-M & 923 & 21 & 48 & 8 & -2.70 & [-4.30, -1.20] \\
SL-N & GPT-N & 901 & 43 & 36 & 20 & +0.70 & [-1.00, +2.40] \\
SL-N & JEV & 905 & 39 & 33 & 23 & +0.60 & [-1.00, +2.20] \\
SL-N & Q38-F & 758 & 186 & 24 & 32 & +16.20 & [+13.80, +18.70] \\
SL-N & Q38-M & 922 & 22 & 43 & 13 & -2.10 & [-3.70, -0.60] \\
Clef & Q35-F & 756 & 152 & 40 & 52 & +11.20 & [+8.70, +13.70] \\
Clef & Q35-T & 849 & 59 & 77 & 15 & -1.80 & [-4.00, +0.40] \\
Clef & GPT-M & 885 & 23 & 86 & 6 & -6.30 & [-8.20, -4.40] \\
Clef & GPT-N & 871 & 37 & 66 & 26 & -2.90 & [-4.80, -1.00] \\
Clef & JEV & 874 & 34 & 64 & 28 & -3.00 & [-4.90, -1.20] \\
Clef & Q38-F & 737 & 171 & 45 & 47 & +12.60 & [+10.00, +15.30] \\
Clef & Q38-M & 883 & 25 & 82 & 10 & -5.70 & [-7.60, -3.80] \\
Q35-F & Q35-T & 762 & 34 & 164 & 40 & -13.00 & [-15.40, -10.60] \\
Q35-F & GPT-M & 777 & 19 & 194 & 10 & -17.50 & [-19.90, -15.10] \\
Q35-F & GPT-N & 769 & 27 & 168 & 36 & -14.10 & [-16.60, -11.80] \\
Q35-F & JEV & 770 & 26 & 168 & 36 & -14.20 & [-16.60, -11.90] \\
Q35-F & Q38-F & 707 & 89 & 75 & 129 & +1.40 & [-1.00, +3.80] \\
Q35-F & Q38-M & 778 & 18 & 187 & 17 & -16.90 & [-19.30, -14.50] \\
Q35-T & GPT-M & 906 & 20 & 65 & 9 & -4.50 & [-6.20, -2.70] \\
Q35-T & GPT-N & 878 & 48 & 59 & 15 & -1.10 & [-3.00, +0.90] \\
Q35-T & JEV & 880 & 46 & 58 & 16 & -1.20 & [-3.20, +0.80] \\
Q35-T & Q38-F & 741 & 185 & 41 & 33 & +14.40 & [+11.80, +17.10] \\
Q35-T & Q38-M & 913 & 13 & 52 & 22 & -3.90 & [-5.50, -2.30] \\
GPT-M & GPT-N & 925 & 46 & 12 & 17 & +3.40 & [+2.00, +4.90] \\
GPT-M & JEV & 919 & 52 & 19 & 10 & +3.30 & [+1.70, +4.90] \\
GPT-M & Q38-F & 763 & 208 & 19 & 10 & +18.90 & [+16.50, +21.40] \\
GPT-M & Q38-M & 948 & 23 & 17 & 12 & +0.60 & [-0.60, +1.80] \\
GPT-N & JEV & 896 & 41 & 42 & 21 & -0.10 & [-1.90, +1.60] \\
GPT-N & Q38-F & 757 & 180 & 25 & 38 & +15.50 & [+13.00, +18.00] \\
GPT-N & Q38-M & 917 & 20 & 48 & 15 & -2.80 & [-4.40, -1.20] \\
JEV & Q38-F & 755 & 183 & 27 & 35 & +15.60 & [+13.20, +18.10] \\
JEV & Q38-M & 917 & 21 & 48 & 14 & -2.70 & [-4.30, -1.10] \\
Q38-F & Q38-M & 769 & 13 & 196 & 22 & -18.30 & [-20.80, -16.00] \\
\end{longtable}

\endgroup

\end{document}